\PassOptionsToPackage{compatibility=false}{caption}
\documentclass[runningheads]{llncs}

\usepackage[T1]{fontenc}
\usepackage{graphicx}
\usepackage{float}
\usepackage{amsmath}
\usepackage{amssymb}
\usepackage{mathtools}
\usepackage{booktabs}
\usepackage{makecell}
\usepackage{array}
\usepackage{threeparttable}
\usepackage{enumitem}
\usepackage{textcomp}
\usepackage{xcolor}
\usepackage{colortbl}
\usepackage{multirow}
\usepackage{silence}
\usepackage[compatibility=false]{caption}
\usepackage{subcaption}
\usepackage{soul}
\usepackage{algorithm}
\usepackage{algpseudocode}
\usepackage{listings}
\usepackage{ragged2e}
\usepackage{url}
\usepackage[breaklinks=true,colorlinks,bookmarks=false]{hyperref}
\usepackage[noabbrev,nameinlink,capitalise]{cleveref}

\makeatletter
\@ifundefined{doi}{\newcommand{\doi}[1]{\url{https://doi.org/#1}}}{\renewcommand{\doi}[1]{\url{https://doi.org/#1}}}
\makeatother
\Crefformat{figure}{#2Fig.~#1#3}
\Crefmultiformat{figure}{Figs.~#2#1#3}{ and~#2#1#3}{, #2#1#3}{ and~#2#1#3}

\sethlcolor{yellow}
\soulregister{\cite}{7}
\soulregister{\ref}{7}
\soulregister{\cref}{7}

\title{A Multimodal Large Language Model-Driven Framework for Context-Aware UAV Emergency Landing Site Selection}
\titlerunning{Context-Aware UAV Emergency Landing Site Selection}
\author{Chunliang HUA$^{1,*}$ \and Lei ZHANG$^{2}$ \and Jiayang SUN$^{2}$\\
Chunlan ZENG$^{3}$ \and Xiao HU$^{2,\dagger}$}
\authorrunning{C. Hua et al.}
\institute{}

\begin{document}
\maketitle
\begingroup
\renewcommand{\thefootnote}{*}
\footnotetext[0]{This work was completed during Chunliang HUA's internship at LASER IDEA.}
\renewcommand{\thefootnote}{\ensuremath{\dagger}}
\footnotetext[0]{Corresponding author: Xiao HU.}
\endgroup
\footnotetext[1]{School of Information Science and Engineering, Southeast University, Nanjing 211189, China (e-mail: chunlianghua@seu.edu.cn).}
\footnotetext[2]{Lower AirSpace Economy Research Institute, International Digital Economy Academy, Shenzhen 510085, China (e-mail: huxiao1@idea.edu.cn, zhanglei@idea.edu.cn, sunjiayang@idea.edu.cn).}
\footnotetext[3]{Department of Electronic Engineering, East China Normal University, Shanghai 200241, China (e-mail: 52285904003@stu.ecnu.edu.cn).}

\begin{abstract}
Safe UAV emergency landing requires more than just identifying flat terrain; it demands understanding complex semantic risks (e.g., crowds, temporary structures) invisible to traditional geometric sensors. 
In this paper, we propose a novel framework leveraging Remote Sensing (RS) imagery and Multimodal Large Language Models (MLLMs) for global context-aware landing site assessment. 
Unlike local geometric methods, our approach employs a coarse-to-fine pipeline: first, a lightweight semantic segmentation module efficiently pre-screens candidate areas; second, a vision-language reasoning agent fuses visual features with Point-of-Interest (POI) data to detect subtle hazards. 
To validate this approach, we construct and release the \textbf{Emergency Landing Site Selection (ELSS)} benchmark at \href{https://github.com/chunlianghua/ELSS-dataset}{\texttt{chunlianghua/ELSS-dataset}}.
ELSS includes a remote-sensing subset for simulation validation and a UAV aerial-video subset for real-world testing. 
Experiments demonstrate that our framework significantly outperforms geometric baselines in risk identification accuracy. Furthermore, qualitative results confirm its ability to generate human-like, interpretable justifications, enhancing trust in automated decision-making.

\end{abstract}

\section{Introduction}
\label{sec:intro}
\begin{figure}[!htbp]  
    \centering  
    \includegraphics[width=1\textwidth]{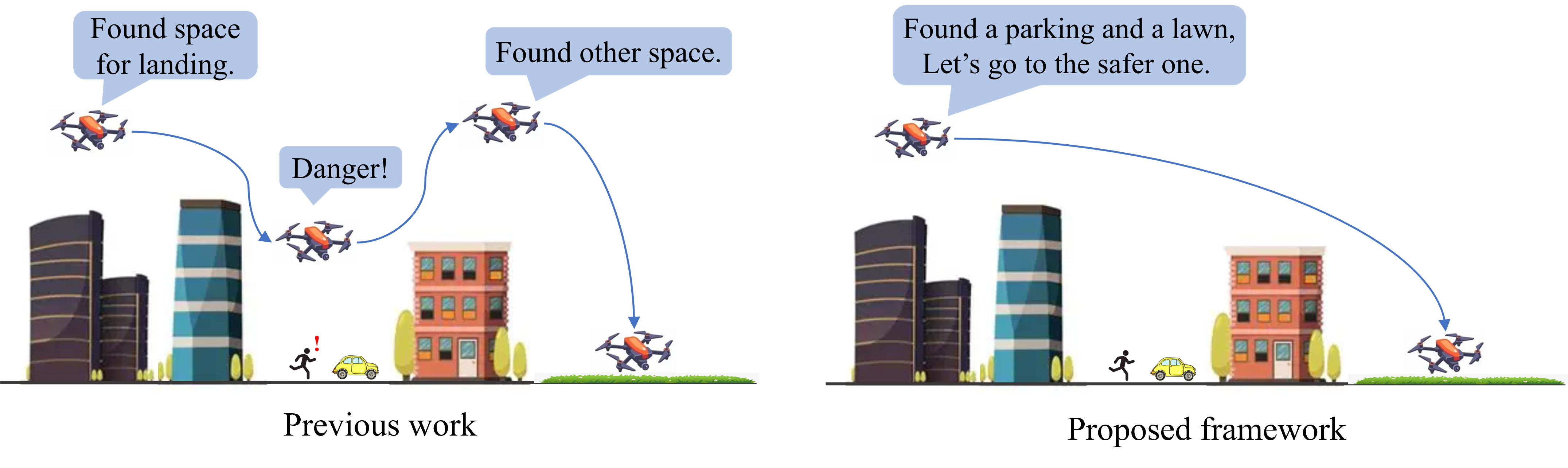}  
\caption{Conceptual comparison between conventional geometry-based approaches and the proposed MLLM-driven framework for emergency landing site selection.}
    \label{fig1-1}  
\end{figure}
Ensuring safety during emergency landings is a critical challenge for Uncrewed Aerial Vehicles (UAVs), particularly in complex urban environments where mechanical failures can lead to catastrophic ground collisions~\cite{Kenny,Australian}. 
Existing solutions primarily rely on geometric evaluation using onboard sensors, such as LiDAR or depth cameras, to identify flat surfaces for landing. These morphology-based methods~\cite{Sujit-icuas2015,Johnson-ICRA2005,Ayhan-TAES-landing,Konig-icuas2025} assess suitability by quantifying slopes and roughness from digital elevation models (DEMs) or point clouds~\cite{Saldiran-TAES-2025,Jeonggeun-TMECH-2024,Scherer2012,Proenca-ICRA2022}.
While effective for local obstacle avoidance, they fundamentally lack \textbf{semantic awareness}. 
In real-world scenarios, geometric flatness does not equate to safety; a flat lawn may be crowded with people, and a smooth parking lot may be an active traffic zone. 
These ``semantic risks'' are invisible to depth sensors but are critical for safe decision-making.

To bridge this gap, recent research in Remote Sensing (RS) has leveraged deep learning for semantic understanding. While domain-specific models like RSGPT~\cite{RSGPT}, GeoChat~\cite{GeoChat}, and EarthGPT~\cite{EarthGPT} demonstrate that terrain semantics significantly enhance interpretation beyond elevation data, applying them to safety-critical landing tasks remains underexplored.
Recent frameworks such as FlightGPT~\cite{cai2025flightgpt} and LLM-Land~\cite{cai2025llmlandlargelanguagemodels} have attempted to use MLLMs for context-aware decisions. However, these approaches often prioritize general autonomy over strict safety compliance (e.g., JARUS SORA guidelines~\cite{jarus}), and they typically lack the explainability required for high-stakes operations.

In this paper, we propose a novel \textbf{Semantic Risk Assessment Framework} that leverages RS imagery and MLLMs to identify safe landing sites from a global perspective. 
Unlike resource-constrained onboard processing, our approach utilizes satellite imagery for macro-scale analysis, serving as a high-level decision support system. 
We introduce a \textit{coarse-to-fine} pipeline: first, a semantic segmentation module filters out obvious non-landable areas; second, a vision-language reasoning agent evaluates candidate sites by fusing visual features with Point-of-Interest (POI) data. 
This allows the system to not only select safe sites but also provide \textbf{interpretable natural language justifications} for its decisions.

The main contributions of this work are summarized as follows: 1) We propose the first MLLM-driven framework for UAV landing site assessment, shifting the paradigm from purely geometric analysis to semantic understanding; 2) We construct and release the \textbf{Emergency Landing Site Selection (ELSS)} benchmark dataset at \href{https://github.com/chunlianghua/ELSS-dataset}{\texttt{chunlianghua/ELSS-dataset}}, which contains diverse urban and rural scenarios with expert annotations to facilitate future research. Experimental results demonstrate that our method significantly outperforms geometric baselines in risk identification. Furthermore, qualitative analysis confirms that our model generates human-like explanations, enhancing the transparency and trustworthiness of the automated system.

The remainder of this paper is organized as follows: Section~\ref{method} details our three-stage pipeline; Section~\ref{exp} presents experimental results; and Section~\ref{conclusion} concludes with future directions.

\section{Related Work}
\label{related}

\subsection{Landing Site Assessment}
Prior work on UAV landing site assessment mainly follows two directions. Geometry-based methods evaluate slope, roughness, clearance, and terrain uncertainty from DEMs, point clouds, or onboard elevation maps~\cite{Sujit-icuas2015,Johnson-ICRA2005,Ayhan-TAES-landing,Konig-icuas2025,Saldiran-TAES-2025,Scherer2012,Vishnu-RSS14,Proenca-ICRA2022}.
Learning-based methods instead predict suitability from visual or semantic cues, ranging from early hazard classification to CNN-based landing assessment and multi-sensor or 3D perception models~\cite{Guo-ICPR2016,Abdollahzadeh-CRV2022,Jeonggeun-TMECH-2024,Maturana15}. These approaches improve local landing assessment, but they still have difficulty handling richer semantic risk, dynamic context, and open-ended safety constraints in a unified way.

\subsection{MLLMs and Context-Aware Geospatial Reasoning}
Recent MLLM-based studies suggest a more flexible route for context-aware UAV reasoning. FlightGPT, LLM-Land, and related surveys indicate the promise of language-guided aerial autonomy~\cite{TIAN2025103158,cai2025flightgpt,cai2025llmlandlargelanguagemodels}, while remote-sensing MLLMs such as RSGPT, GeoChat, and EarthGPT show that geospatial semantics can be interpreted beyond elevation maps or segmentation labels~\cite{RSGPT,GeoChat,EarthGPT}. However, existing systems focus more on general autonomy or broad remote-sensing understanding than on safety-critical landing decisions under explicit constraints such as JARUS SORA~\cite{jarus}. Our work addresses this gap by combining MLLMs with POI semantics and scene context for interpretable, safety-oriented landing-site assessment.

\section{Method}
\label{method}
\begin{figure}[!htbp]  
    \centering  
    \includegraphics[width=1\textwidth]{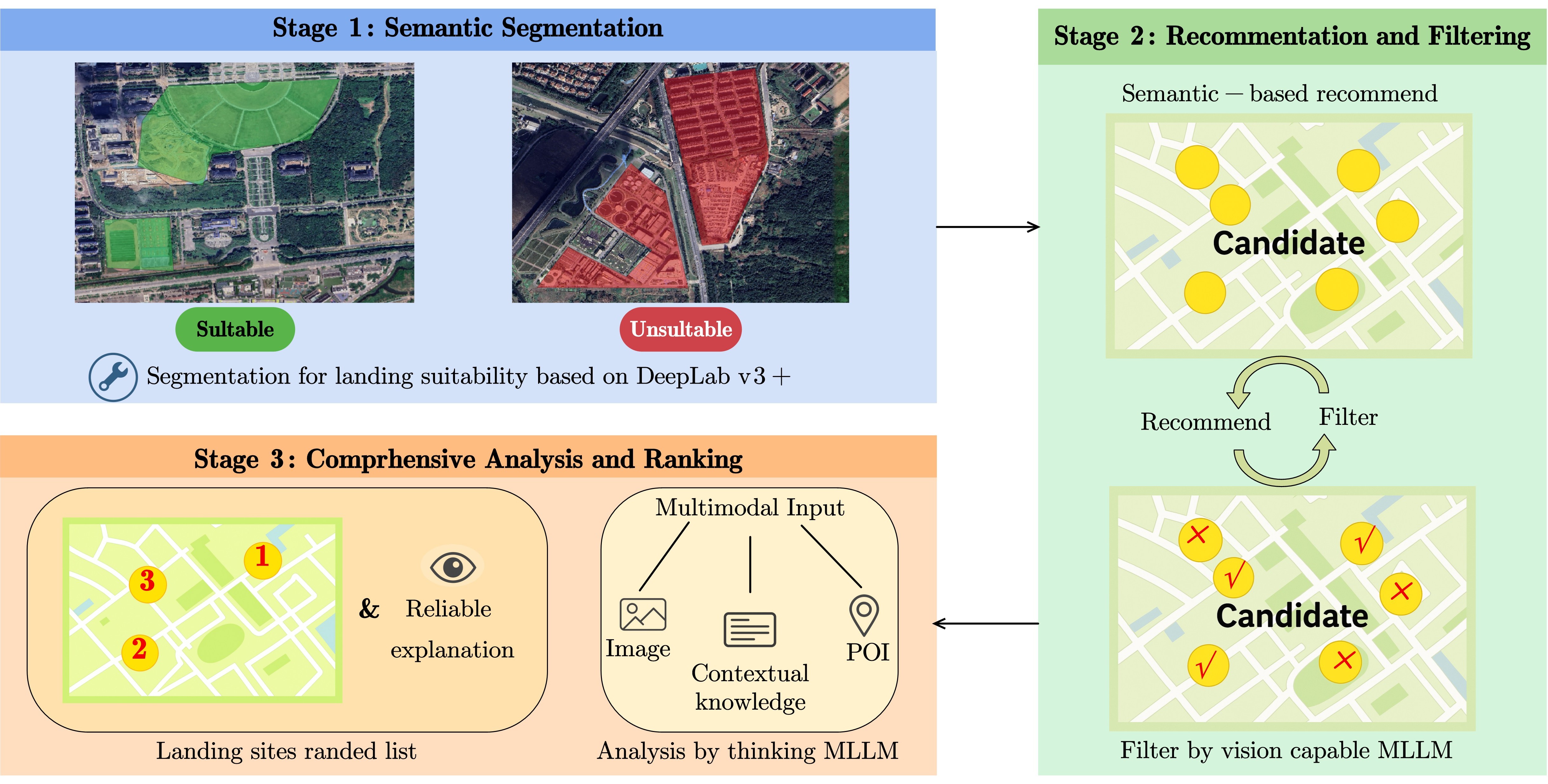}  
    \caption{Scheme chart of the proposed framework.} 
    \label{fig3-1}  
\end{figure}

The proposed framework, illustrated in Fig.~\ref{fig3-1}, is structured as a coarse-to-fine pipeline designed to simulate human pilot cognition. It progressively filters landing candidates by integrating visual perception with semantic reasoning.
\subsection{Problem Formulation}
Let $\mathcal{I} \in \mathbb{R}^{H \times W \times 3}$ be the input remote sensing image and $\mathcal{P} = \{p_1, p_2, ..., p_k\}$ be the set of nearby Point-of-Interest (POI) data. Our goal is to identify a ranked list of landing sites $\mathcal{S} = \{s_1, s_2, ..., s_m\}$, where each site $s_i$ is defined by a bounding box $b_i$, a safety score $\sigma_i \in [0,1]$, and a natural language justification $\mathcal{T}_i$.
The problem is modeled as a hierarchical filtering process:
\begin{equation}
    \mathcal{S} = f_{\text{reason}}(f_{\text{verify}}(f_{\text{seg}}(\mathcal{I})), \mathcal{P}, \mathcal{C}_{dyn})
\end{equation}
where $f_{\text{seg}}$, $f_{\text{verify}}$, and $f_{\text{reason}}$ correspond to the semantic segmentation, visual verification, and context reasoning modules, respectively, and $\mathcal{C}_{dyn}$ represents dynamic context (e.g., time, weather).
\begin{figure}[!htbp]  
    \centering
    \begin{subfigure}[b]{0.3\linewidth}
        \includegraphics[width=\linewidth]{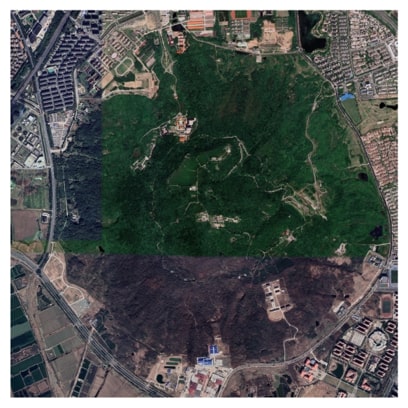}
        \caption{}
        \label{fig3-2-1}
    \end{subfigure}
    \hfill
    \begin{subfigure}[b]{0.3\linewidth}
        \includegraphics[width=\linewidth]{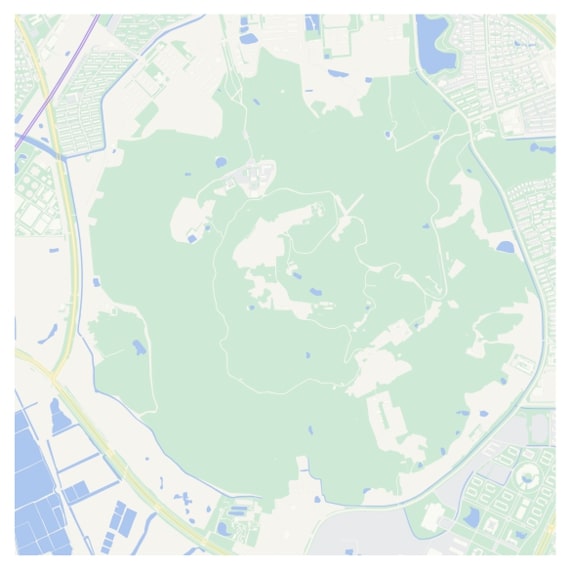}
        \caption{}
        \label{fig3-2-2}
    \end{subfigure}
    \hfill
    \begin{subfigure}[b]{0.3\linewidth}
        \includegraphics[width=\linewidth]{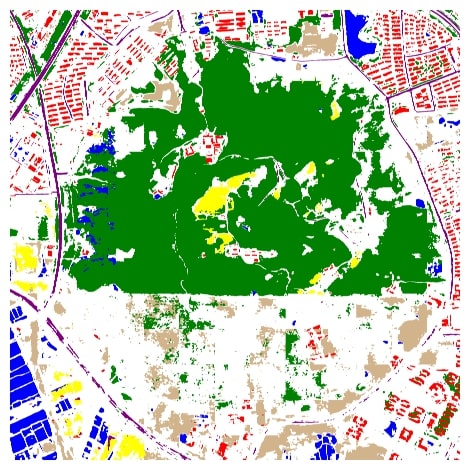}
        \caption{}
        \label{fig3-2-3}
    \end{subfigure}
    \caption{
    Comparison of remote sensing imagery and standard maps highlighting segmentation results. The figure shows correct segmentation of the upper forest half and misclassification of the lower half as background, emphasizing the limitations of using only satellite images for segmentation.
    }
    \label{fig3-2}
\end{figure}
\subsection{Stage 1: Semantic Filtering via Segmentation}
\label{sec:stage1}
The first stage aims to efficiently parse large-scale remote sensing imagery to generate a set of potential landing candidates, reducing the search space for the subsequent MLLM. We employ the DeepLabV3+~\cite{DeepLab} framework (implemented via MMSegmentation~\cite{MMSegmentation}) due to its superior capability in handling scale variations inherent in aerial views. Unlike standard U-Net architectures, DeepLabV3+ utilizes Atrous Spatial Pyramid Pooling (ASPP) to capture multi-scale context without losing resolution. 
Mathematically, the atrous convolution for a signal $x$, filter $w$, and dilation rate $r$ is defined as $y[i] = \sum_k x[i + r \cdot k] \cdot w[k]$. 
This mechanism allows the network to distinguish between large homogeneous regions (e.g., lakes) and fine-grained structures (e.g., narrow roads), which is critical for identifying safe landing zones.

\begin{figure}[!htbp]
  \centering
  \includegraphics[width=0.65\linewidth]{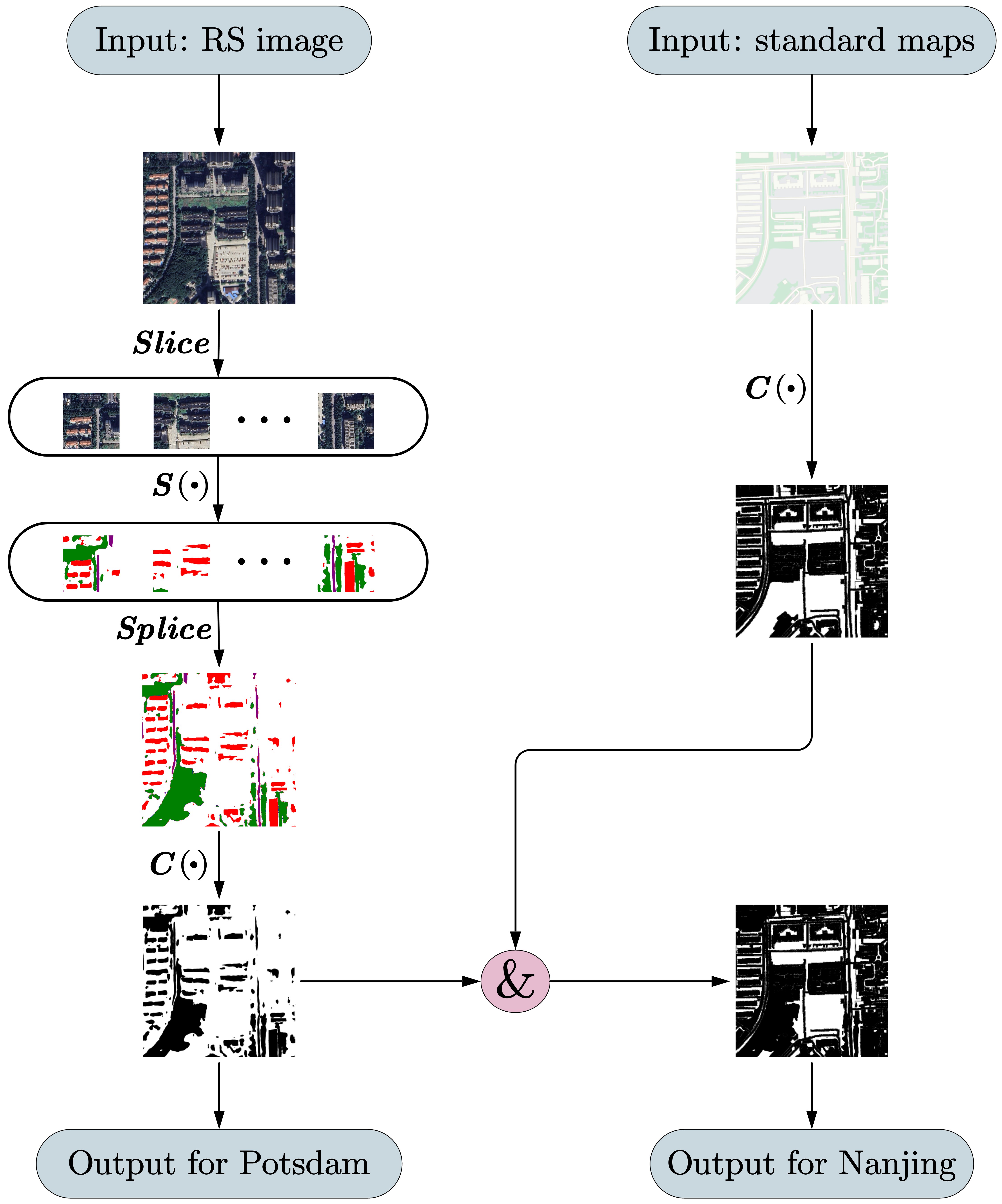}
  \caption{The flow chart of proposed stage 1. Specifically, \( S(\cdot) \) denotes the semantic segmentation network, and \( C(\cdot) \) corresponds to the classification head that evaluates pixel-wise landing suitability.
}
  \label{fig3-3}
\end{figure}

However, deploying this model reveals significant \textit{domain shift} challenges. While the network achieves high accuracy on high-resolution aerial datasets like ISPRS Potsdam~\cite{Potsdam}, its performance degrades on lower-resolution satellite imagery (e.g., our Nanjing dataset, see Sec.~\ref{exp}). As illustrated in Fig.~\ref{fig3-2}, differences in texture and spectral characteristics lead to systematic misclassifications, such as dense forests being labeled as soil due to texture smoothing. Relying solely on visual segmentation in such heterogeneous data sources risks proposing hazardous sites. To address these misclassification issues, a strategy was proposed, as described in \Cref{fig3-3}, to cross-validate the classification results using map information. This flow chart illustrates the integration of a semantic segmentation network, \( S(\cdot) \), with a classification head, \( C(\cdot) \), to evaluate pixel-wise landing suitability. By leveraging standard maps, this method enhances the reliability of pixel classification and addresses domain adaptation challenges. Ultimately, the semantic output not only identifies potential candidate sites for further analysis but also enhances efficiency by acting as a crucial filtering step. This hybrid approach transforms the segmentation step from a deterministic classifier into a \textit{high-recall candidate generator}, ensuring that only plausibly safe regions are passed to the MLLM for fine-grained verification.

\subsection{Stage 2: Visual Verification via Iterative Proposals}
\label{sec:stage2}
\begin{figure}[!htbp]
  \centering
  \includegraphics[width=0.6\linewidth]{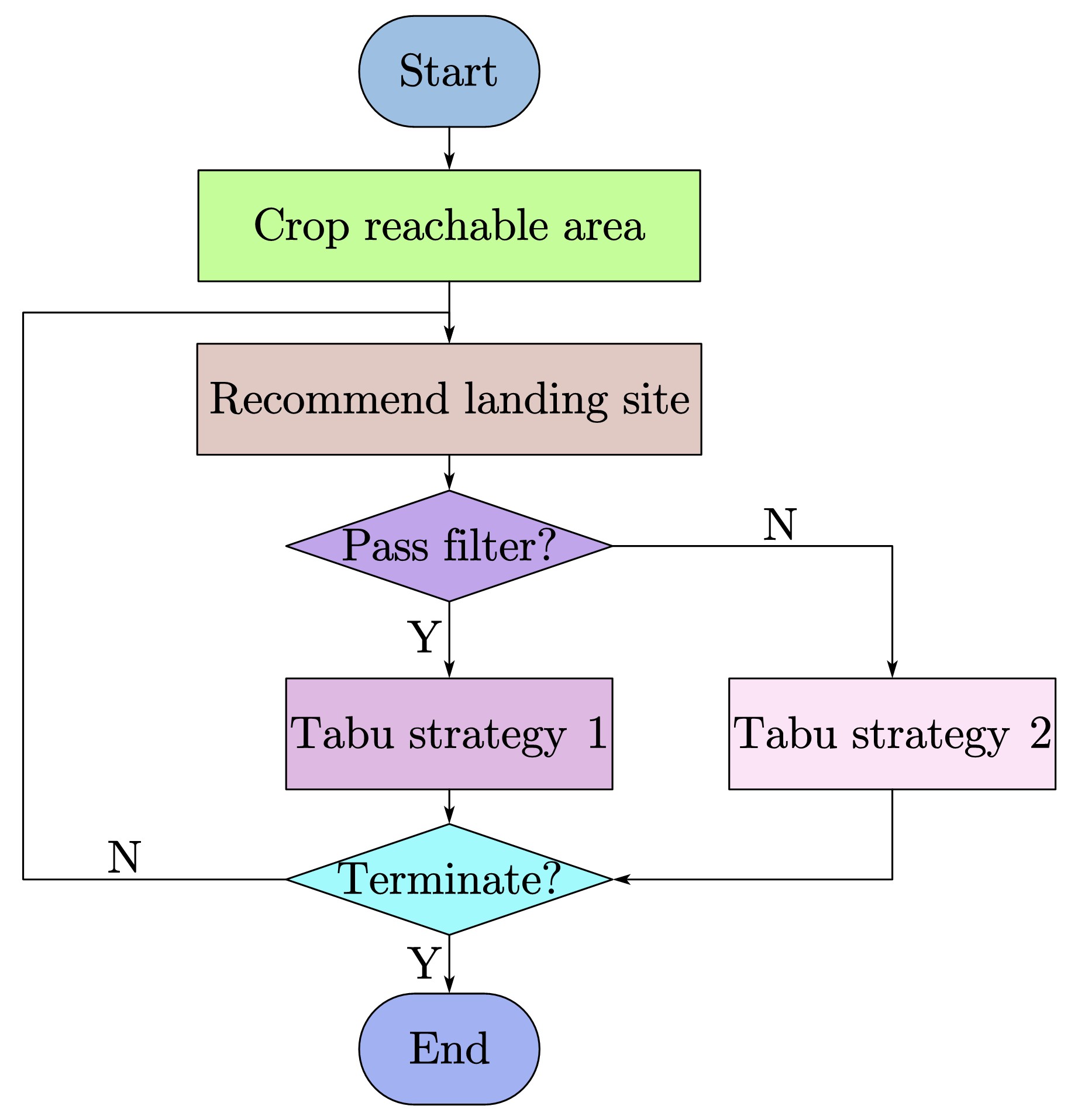}
  \caption{Flow chart of the recommendation and filtering scheme.}
  \label{fig3-4}
\end{figure}

While semantic segmentation offers a global view of land use, it lacks the fine-grained reasoning to distinguish between a ``flat plaza'' and a ``cracked pavement.'' To address this, we implement an iterative \textit{propose-and-verify} loop (Fig.~\ref{fig3-4}) that bridges pixel-level cues with high-level semantic reasoning.

The process begins by formalizing candidate generation as a spatial density search. We define a radially decaying convolution kernel $K$ to identify centers of large, contiguous suitable regions. The kernel weights are designed to favor centrality:
\begin{equation}
    K_{i,j} = 1 - \frac{(i - d)^2 + (j - d)^2}{2d^2}, \ \text{for } i,j \in \{0, \dots, 2d\}
\end{equation}
A response map is computed via $R = I * K$, where $I$ is the binary suitability map from Stage 1. The location of the maximum response $(x^*, y^*) = \arg\max R(x,y)$ serves as the center for the current landing proposal. This convolution-based approach naturally filters out fragmented noise and prioritizes compact regions suitable for the UAV's footprint.

Once a candidate is proposed, we crop the corresponding high-resolution image patch and query a vision-capable VLM to detect hazards invisible to the segmentation model (e.g., vehicles, rubble). The model is instructed via a specific prompt: \textit{``Inspect this satellite image patch. Is the surface smooth and free of dynamic obstacles or structural hazards? Answer [Safe/Unsafe] and provide a 1-sentence reason.''}
Based on the VLM's verdict, we employ a dual-strategy \textbf{Tabu Mechanism} to dynamically update the response map $R(x,y)$ for the next iteration:
\begin{itemize}[leftmargin=*]
    \item \textbf{Successful Filter (Hard Suppression):} If the site is confirmed as Safe, we zero out the local neighborhood $\mathcal{N}$ to enforce diversity: $R'(x, y) = 0, \forall (x, y) \in \mathcal{N}(x^*, y^*)$.
    \item \textbf{Failed Filter (Soft Penalty):} If rejected (e.g., due to temporary occupancy), we apply a Gaussian decay penalty to discourage revisiting the exact spot while allowing valid candidates nearby:
    \begin{equation}
        R'(x, y) = R(x, y) \cdot \frac{(x - x^*)^2 + (y - y^*)^2}{2d^2}, \forall (x, y) \in \mathcal{N}
    \end{equation}
\end{itemize}
This feedback loop continues iteratively, adaptively guiding the search toward novel, high-quality candidates without exhaustive re-sampling.

\subsection{Stage 3: Context-Aware Reasoning via Multi-Modal Fusion}
\label{sec:stage3}

\begin{figure}[!htbp]  
    \centering  
    \includegraphics[width=0.65\linewidth]{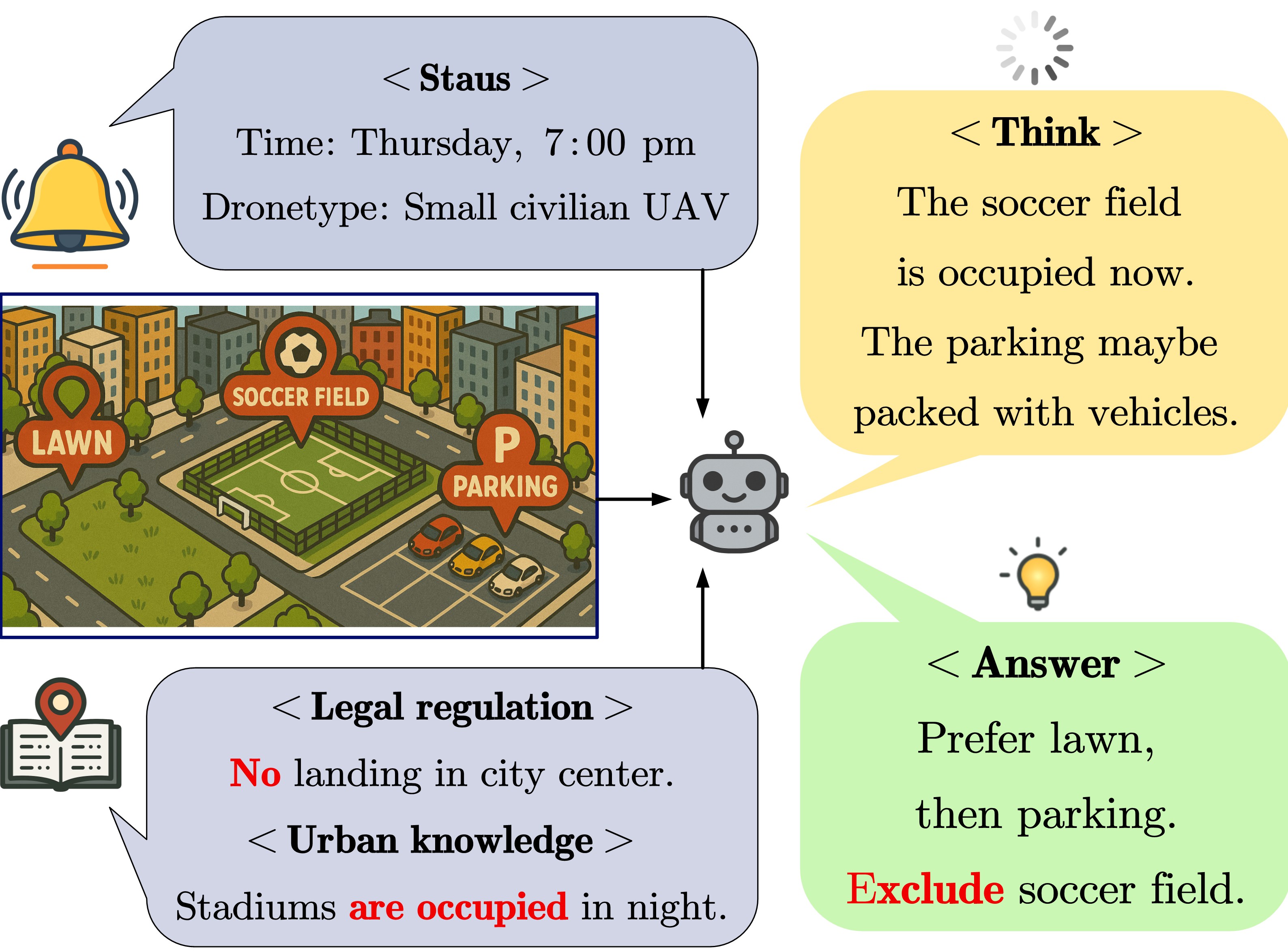}  
    \caption{Flowchart of the MLLM-driven comprehensive analysis stage, integrating remote sensing, POI, dynamic context, and domain knowledge for final site ranking.} 
    \label{fig3-5}  
\end{figure}

The final stage mimics the high-level cognitive process of a human pilot, selecting the optimal landing site not just based on geometry, but on comprehensive safety protocols. We employ a reasoning-oriented MLLM to synthesize four heterogeneous information streams (Fig.~\ref{fig3-5}):
\begin{itemize}[leftmargin=*]
    \item \textbf{Visual Evidence:} The high-resolution cropped patch of the candidate site passed from Stage 2.
    \item \textbf{Spatial Context (POI):} Static vector data retrieved from OpenStreetMap or Amap, quantifying proximity to sensitive facilities (e.g., schools, gas stations, power lines).
    \item \textbf{Dynamic Context:} Real-time environmental variables, including time of day, weather conditions, and active social events (e.g., ``Rush Hour'', ``Public Holiday'').
    \item \textbf{Regulatory Constraints:} A system prompt injecting domain-specific rules derived from JARUS SORA guidelines (e.g., ``Maintain 1:1 buffer distance from people'').
\end{itemize}

By integrating these inputs, the MLLM performs \textit{spatio-temporal reasoning}. For instance, it recognizes that a school playground, while visually flat (Safe geometrically), is a high-risk zone during weekdays at 10:00 AM (Unsafe semantically) but potentially viable at midnight. 
The model outputs a strictly ranked list of sites, where each recommendation is accompanied by a \textbf{Natural Language Justification} (e.g., \textit{``Rank 1: Empty parking lot. Justification: Surface is flat, far from high-rise buildings, and currently outside of business hours with low dynamic risk.''}). This interpretability is crucial for auditability in safety-critical operations.

\section{Experimental Results and Analysis}
\label{exp}

\begin{figure}[!htbp]
  \centering
  \begin{subfigure}{0.45\linewidth}
    \centering
    \includegraphics[width=1.0\linewidth]{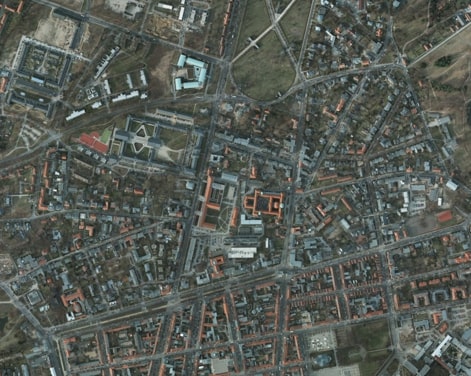}
    \caption{\centering The Potsdam dataset.}
    \label{fig4-1}
  \end{subfigure}
  \begin{subfigure}{0.45\linewidth}
    \centering
    \includegraphics[width=\linewidth]{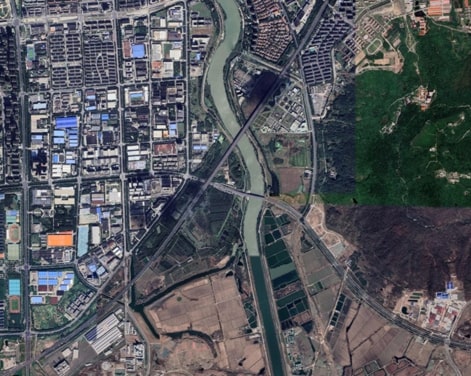}
    \caption{\centering The Nanjing dataset.}
    \label{fig4-2}
  \end{subfigure}
  \caption{Comparison of datasets.}
  \label{fig:datasets}
\end{figure}

\subsection{Study Area and Datasets}

We evaluate our framework on two publicly available remote sensing datasets with distinct acquisition platforms: Potsdam (airborne) and Nanjing (satellite):
\begin{itemize}[leftmargin=*]
    \item The Potsdam dataset~\cite{Potsdam} consists of 38 aerial images, each with a resolution of 0.05 m and dimensions of 6000×6000 pixels, depicting a dense urban area in Germany. We utilized 20 tiles to create a composite image measuring 24,000×30,000 pixels (see~\cref{fig4-1}). Segmentation is performed using the DeepLab v3+ model\footnote{Model version: deeplabv3plus\_r50-d8\_4xb4-80k\_Potsdam-512x512.} pretrained on Potsdam, which outputs six classes; we treat Background and Impervious Surface as landing-suitable.
    \item The Nanjing dataset, derived from Google Earth, covers a broader urban–rural fringe in China at 0.3 m resolution of 15,000×30,000 pixels, see~\cref{fig4-2} as an example. Despite lower resolution, it spans over 20× the ground area of Potsdam and includes forests, farmland, water, and barren land. As no DeepLab v3+ model is directly trained on this dataset, we utilized the model\footnote{Model version: deeplabv3plus\_r50-d8\_4xb4-80k\_loveda-512x512.} pretrained on the LoveDA dataset, which incorporates Nanjing samples. This model predicts seven classes, with Background and Barren classified as suitable for landing.
\end{itemize}
These datasets represent contrasting regimes: high-resolution, costly airborne imagery (Potsdam) vs. freely available, lower-resolution satellite data (Nanjing). This difference necessitates minor adaptations in segmentation and POI sourcing: OSM for Potsdam, Amap for Nanjing.

\begin{figure}[!htbp]
  \centering
  \includegraphics[width=0.9\linewidth]{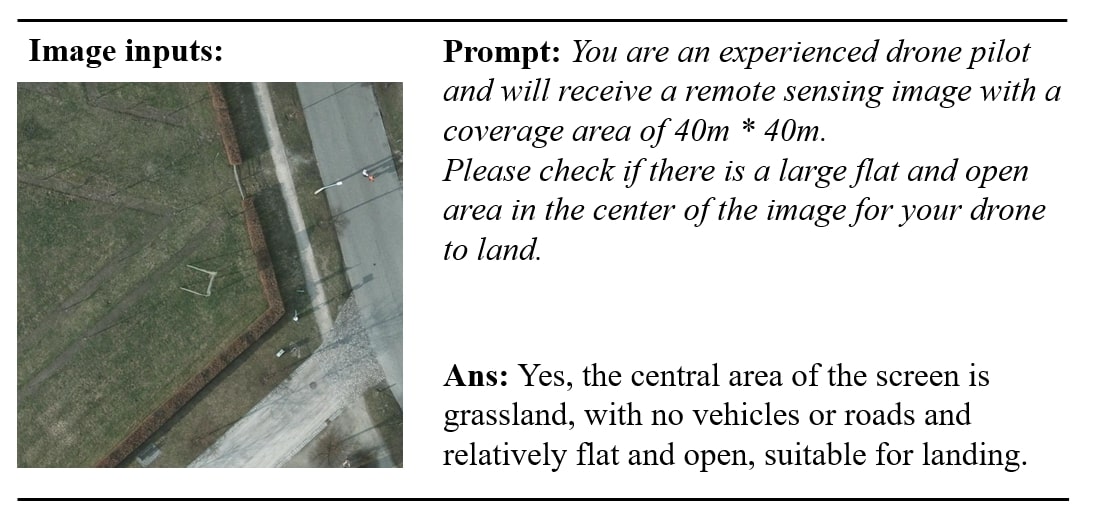}
  \caption{MLLM invocation workflow for filtering candidate landing sites.}
  \label{fig4-3}
\end{figure}

\subsection{Analysis of Candidate Site Filtering}

\begin{table}[!htbp]
\centering
\caption{Proportion of candidate sites passing the MLLM-based filter, by dataset and recommendation method.}
\label{table1}
\begingroup
\footnotesize
\setlength{\tabcolsep}{8pt}
\renewcommand{\arraystretch}{1.08}
\begin{tabular}{@{}lcccc@{}}
\Xhline{1.2pt}
\multirow{2}{*}{\textbf{MLLM Model}} 
& \multicolumn{2}{c}{\textbf{Potsdam}} 
& \multicolumn{2}{c}{\textbf{Nanjing}} \\
\cmidrule(r){2-3} \cmidrule(l){4-5}
& \textbf{Random} & \textbf{Semantic} 
& \textbf{Random} & \textbf{Semantic} \\
\midrule
GPT    & 12 & 44\,\textcolor{red}{$\uparrow$} & 12 & 51\,\textcolor{red}{$\uparrow$} \\
Gemini & 9  & 43\,\textcolor{red}{$\uparrow$} & 13 & 44\,\textcolor{red}{$\uparrow$} \\
Doubao & 14 & 53\,\textcolor{red}{$\uparrow$} & 19 & 51\,\textcolor{red}{$\uparrow$} \\
Claude & 3  & 30\,\textcolor{red}{$\uparrow$} & 3  & 48\,\textcolor{red}{$\uparrow$} \\
Grok   & 15 & 53\,\textcolor{red}{$\uparrow$} & 24 & 62\,\textcolor{red}{$\uparrow$} \\
\Xhline{1.2pt}
\end{tabular}
\endgroup
\end{table}

We evaluate five recent high-performing vision-capable MLLMs selected according to the Chatbot Arena Vision leaderboard\footnote{\url{https://openlm.ai/chatbot-arena/}}: \textit{GPT-5.4} (GPT), \textit{Gemini-3.1-Pro} (Gemini), \textit{Doubao-Seed-2.0-Pro} (Doubao), \textit{Claude Opus 4.6} (Claude), and \textit{Grok-4.20} (Grok). Table~\ref{table1} reports the proportion of candidate sites passing our MLLM-based visual filter under random and semantic-based recommendation strategies. We keep this comparison as an efficiency-oriented sanity check, but the gains are still substantial. On Potsdam, the pass rate rises from 3--15\% under random sampling to 30--53\% under semantic-based recommendation. A similar trend appears on Nanjing, where the pass rate rises from 12--24\% under random sampling to 44--62\% under semantic-based recommendation. These consistent improvements show that semantic pre-screening remains necessary even when the downstream visual filter is implemented by stronger MLLMs, because it reduces the number of unsuitable regions sent to costly MLLM inference.

We then manually annotate semantic-based recommendations and evaluate the filtering quality using precision (Prec.), recall (Rec.), and F1 score, as summarized in~\cref{table2}. The human-positive ratios are 57\% for Potsdam and 63\% for Nanjing, indicating that these recommended candidates still contain a substantial number of unsuitable sites. Therefore, Stage II is not merely a confirmation step; it is required to remove false positives before final contextual ranking.

The models also exhibit distinct filtering styles. GPT shows the largest cross-dataset variance: it performs strongly on Nanjing, but noticeably worse on Potsdam, mainly because it sometimes overlooks small shrubs or dried branches and occasionally over-penalizes narrow paved paths in grassy areas. Claude is the most conservative, reaching the highest Potsdam precision (96.7\%) at the cost of the lowest recall (50.9\%). Gemini shows a milder version of the same high-precision, low-recall tendency, whereas Grok is more permissive and achieves the best Nanjing F1 score through higher recall. Overall, Doubao achieves the highest mean F1 score across the two datasets and offers the most balanced trade-off between precision and recall. We therefore select Doubao as the MLLM used in the subsequent landing-site ranking experiments.

\begin{table}[!htbp]
\centering
\caption{Precision, recall, and F1 score of the MLLM-based filter on semantic-based recommendations. Rows are sorted by the mean F1 score across the two datasets.}
\label{table2}
\begingroup
\footnotesize
\setlength{\tabcolsep}{7pt}
\renewcommand{\arraystretch}{1.08}
\makebox[\dimexpr\width+4em\relax][c]{%
\begin{tabular}{lcccccc}
\Xhline{1.2pt}
\multirow{2}{*}{\textbf{MLLM Model}} 
& \multicolumn{3}{c}{\textbf{Potsdam}} 
& \multicolumn{3}{c}{\textbf{Nanjing}} \\
\cmidrule(r){2-4} \cmidrule(l){5-7}
& \textbf{Prec.} & \textbf{Rec.} & \textbf{F1} 
& \textbf{Prec.} & \textbf{Rec.} & \textbf{F1} \\
\midrule
Doubao & 83.0 & \textcolor{red}{77.2} & \textcolor{red}{80.0} & 86.3 & 69.8 & 77.2 \\
Grok   & 75.5 & 70.2 & 72.7 & 83.9 & \textcolor{red}{82.5} & \textcolor{red}{83.2} \\
GPT    & 75.0 & 57.9 & 65.3 & \textcolor{red}{92.2} & 74.6 & 82.5 \\
Gemini & 86.0 & 64.9 & 74.0 & 88.6 & 61.9 & 72.9 \\
Claude & \textcolor{red}{96.7} & 50.9 & 66.7 & 83.3 & 63.5 & 72.1 \\
\Xhline{1.2pt}
\end{tabular}%
}
\endgroup
\end{table}

\subsection{Analysis of Landing Site Ranking and Evaluation}

\begin{figure}[!htbp]
  \centering
  \includegraphics[width=0.9\linewidth]{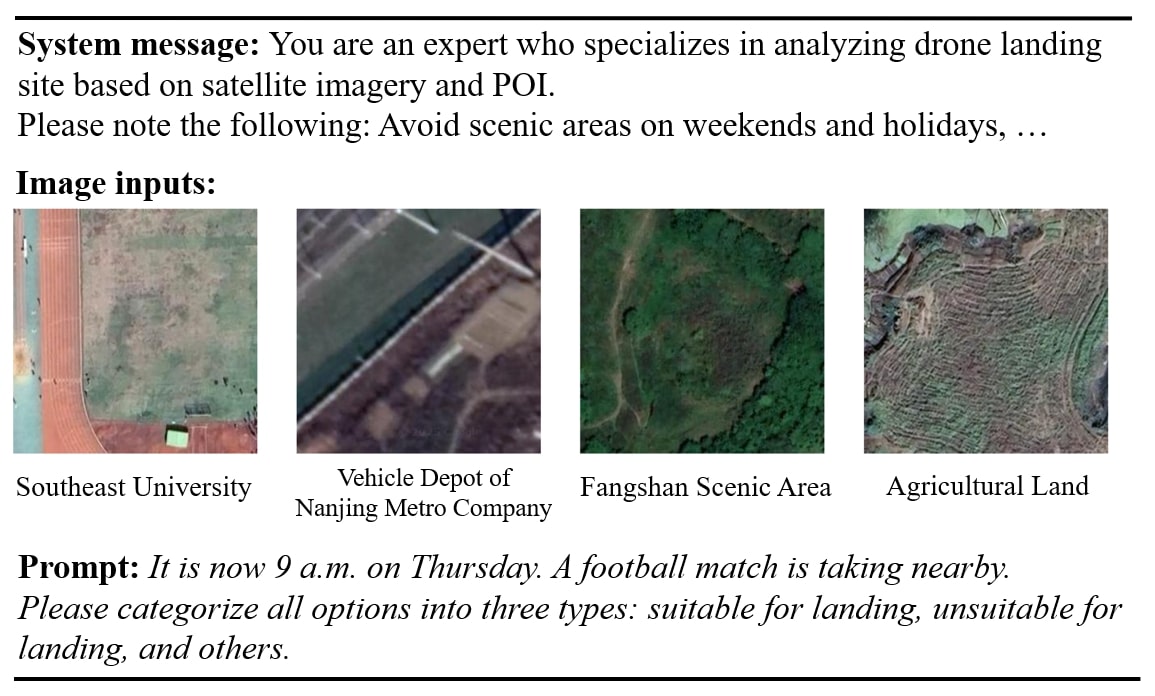}
  \caption{Sample query from the custom benchmark: four candidate landing sites, each with RS imagery and POI context, labeled for suitability classification.}
  \label{fig4-8}
\end{figure}

To evaluate the third stage of our framework, we constructed a curated benchmark consisting of 200 manually annotated queries (100 per dataset). Each query presents four candidate landing sites, each represented by a remote sensing image and associated POI data (Fig.~\ref{fig4-8}). For each set, annotators identified one most suitable and one least suitable site; the model must classify all four into one of three categories: suitable, unsuitable, or other. We use Doubao as the baseline MLLM due to its strong filtering performance in Stage II. Results are summarized in~\cref{table3}, where:
\begin{itemize}[leftmargin=*]
\item \textbf{Right Rate (R.R.)}: Queries where the best site is classified as suitable and the worst site as unsuitable.
\item \textbf{False Rate (F.R.)}: Queries where the best site is classified as unsuitable, or the worst site as suitable.
\item \textbf{Other}: All remaining cases without a clear correct/false outcome.
\end{itemize}

\begin{table}[htbp]
\centering
\caption{Landing site ranking performance on the custom benchmark.}
\label{table3}
\begingroup
\footnotesize
\setlength{\tabcolsep}{7pt}
\renewcommand{\arraystretch}{1.08}
\begin{tabular}{lcccccc}
\Xhline{1.2pt}
\multirow{2}{*}{\textbf{Method}} 
& \multicolumn{3}{c}{\textbf{Potsdam}} 
& \multicolumn{3}{c}{\textbf{Nanjing}} \\
\cmidrule(r){2-4} \cmidrule(l){5-7}
& \textbf{R.R.} & \textbf{F.R.} & \textbf{Other} 
& \textbf{R.R.} & \textbf{F.R.} & \textbf{Other} \\
\midrule
Proposed method & \textcolor{red}{68} & 11 & 21 & \textcolor{red}{67} & 8  & 25 \\
MLLM w/o POI    & $40^{\textcolor{blue}{28\downarrow}}$ & 28 & 32 & $44^{\textcolor{blue}{23\downarrow}}$ & 27 & 29 \\
Random          & 11 & {33} & {56} & 11 & {33} & {56} \\
\Xhline{1.2pt}
\end{tabular}
\endgroup
\end{table}

The results indicate that the proposed method achieves a \textbf{R.R.} of 68\% (Potsdam) and 67\% (Nanjing), demonstrating effective fusion of terrain semantics, POI context, and dynamic cues. In ablation studies without POI input, accuracy drops by \textit{28\% and 23\%}, respectively, confirming that POI integration is critical for reliable reasoning. 

Further investigation showed that most false classifications occurred in complex scenarios involving subtle temporal or contextual cues, such as a prompt stating a square was ``empty at 5 a.m.,'' which the model failed to prioritize. These cases reflect limitations in current MLLM reasoning, not inherent flaws in the framework. When excluding these 80 edge cases, the False Rate falls to just 5\%, indicating that our system achieves high reliability under typical operational conditions. Importantly, as MLLMs evolve through improved reasoning architectures or targeted fine-tuning on domain-specific cues, we expect this performance gap to narrow significantly.

\section{Conclusion and Discussion}
\label{conclusion}

\subsection{Conclusion}
In this work, we presented a novel MLLM-driven framework for UAV emergency landing site assessment, shifting the paradigm from geometric morphology to semantic risk understanding.
By integrating a coarse-to-fine segmentation pipeline with a vision-language reasoning agent, our method effectively identifies semantic hazards (e.g., crowds, temporary structures) invisible to traditional sensors while minimizing computational overhead.
Experiments on the proposed ELSS benchmark demonstrate that our approach significantly outperforms baselines in risk identification accuracy. Crucially, the framework generates human-like, natural language justifications, ensuring the transparency and auditability required for safety-critical operations.

Current limitations include the inference latency of cloud-based MLLMs and the reliance on third-party satellite imagery quality.
Future work will focus on two key directions: (1) \textbf{Model Distillation}, compressing reasoning capabilities into lightweight, edge-optimized networks for offline onboard deployment; and (2) \textbf{Dynamic Fusion}, integrating real-time data feeds (e.g., live traffic, weather) to further enhance situational awareness in complex urban environments.

\subsection{Discussion on Latency and Safety Trade-offs}
For emergency landing, decision latency is part of the safety problem rather than a separate engineering concern. Although cloud MLLMs provide strong semantic reasoning, their response time may limit direct deployment in tightly time-constrained scenarios.

The current framework already addresses part of this issue through coarse-to-fine filtering. A lightweight DeepLabV3+ model first scans the full image and removes clearly unsuitable regions, so MLLM calls are reserved for a much smaller set of difficult candidates. This design reduces average runtime while keeping high-level semantic judgment where it is most useful.

Further reductions may come from two complementary directions. One is parallel state-space evaluation, where multiple short decision branches or candidate combinations are packaged into a batched request instead of being queried strictly in sequence. The other is static context pre-analysis: since RS imagery, POIs, and regulatory constraints are mostly static or slowly changing, an offline regional landing-risk map could be prepared in advance, allowing online decision-making to rely more on fast spatial retrieval and less on full real-time reasoning.

\bibliographystyle{splncs04}
\bibliography{llm,roboconf,cv}

\end{document}